\documentclass[12pt,twoside,reqno]{amsart}

\usepackage[colorlinks=true,citecolor=blue]{hyperref}
\usepackage{mathptmx, amsmath, amssymb, amsfonts, amsthm, mathptmx, enumerate, color, mathrsfs}
\usepackage{graphicx}
\usepackage{booktabs}
\usepackage{subfig}
\usepackage{hyperref}
\usepackage{amsmath}

\usepackage{multirow}
\usepackage{epstopdf}
\usepackage{multicol}
\usepackage{algorithm}
\usepackage{algorithmic}
\usepackage{epstopdf}

\usepackage{cite}

\usepackage{tikz}
\usetikzlibrary[graphs]

\newtheorem{theorem}{Theorem}[section]

\theoremstyle{definition}
\newtheorem{definition}[theorem]{Definition}

\numberwithin{equation}{section}

\DeclareMathOperator*{\argmin}{\arg\!\min}

\renewcommand{\O}[1]{\ensuremath{\mathsf{#1}}} 

\newcommand\independent{\protect\mathpalette{\protect\independenT}{\perp}}
\def\independenT#1#2{\mathrel{\rlap{$#1#2$}\mkern2mu{#1#2}}}

\begin{document}
\setcounter{page}{1}

\vspace*{2.0cm}
\title[Overcoming Representation Bias in Fairness]{Overcoming Representation Bias in Fairness-Aware data Repair using Optimal Transport }
\author[A. Langbridge, A. Quinn, R. Shorten]{Abigail Langbridge$^{1}$, Anthony Quinn$^{1,2,*}$, Robert Shorten$^1$}
\maketitle
\vspace*{-0.6cm}

\begin{center}
{\footnotesize

$^1$Dyson School of Design Engineering, Imperial College London\\
$^2$Department of Electronic and Electrical Engineering, Trinity College Dublin

}\end{center}

\vskip 4mm {\footnotesize \noindent {\bf Abstract.}
Optimal transport (OT) has an important role in transforming data distributions in a manner which engenders fairness. Typically, the OT operators are learnt from the unfair attribute-labelled data, and then used for their repair. Two significant limitations of this approach are as follows: (i) the OT operators for underrepresented subgroups are poorly learnt (i.e. they are susceptible to  representation bias); and (ii) these OT repairs cannot be effected on identically distributed but  out-of-sample (i.e.\ archival) data. In this paper, we address both of these problems by adopting a Bayesian nonparametric stopping rule for learning each attribute-labelled component of the data distribution. The induced OT-optimal quantization operators can then be used to repair the archival data. We formulate a novel definition of the fair distributional target, along with quantifiers that allow us to trade fairness against damage in the transformed data. These are used to  reveal excellent performance of our representation-bias-tolerant scheme in simulated and benchmark data sets.



\renewcommand{\thefootnote}{}
\footnotetext{ $^*$Corresponding author: {\tt a.quinn@ic.ac.uk}

}

\section{Introduction}

Representation bias is a key issue in machine learning, with many classic datasets being biased towards majority groups such as Americans, white people, or men depending on the context \cite{mehrabi2021survey, shankar2017no}. These biases -- when left unaddressed -- can limit trust in model outcomes, undermine efforts towards fairness correction, and exacerbate existing socioeconomic inequalities \cite{shahbazi2023representation}. By designing a fair transformation which directly addresses the issue of representation bias, we can ensure fairness across all population subgroups.


The problem of generalisation is of critical importance in machine learning \cite{mohri2018foundations}. A model's ability to perform well on new, unseen data after being trained on a specific dataset is crucial to facilitate deployment. Over- and under-fitting are common, which lead to underlying causal relationships in data not being captured. This can be as a result of poor modelling choices \cite{mohri2018foundations}, or as a function of the size and composition of training data which can significantly effect the performance of machine learning models \cite{foody1995effect, oates1997effects, shalev2008svm}.

One area where good generalisation is essential is in AI Fairness (AIF). Poor generalization often leads to biased models that perform well on certain segments of the population (typically those overrepresented in the training data) but poorly on others \cite{barocas2023fairness, mehrabi2021survey}. This can result in unfair treatment of underrepresented groups and exacerbate societal biases present in the underlying data.

There are a number of well-established methods for fairness correction which aim to remove or reduce the level of bias in models through changes to the model itself \cite{chzhen2020fair, zafar2017fairness} or the data \cite{salimi2019interventional, feldman2015certifying, gordaliza2019obtaining}. A contemporary review of works in AIF is presented in \cite{barocas2023fairness}. While these data repair methods are popular since they don't restrict downstream model choice, they often require access to the entire (finite and static) dataset in order to conduct their repair - i.e. they are not designed to generalise. The previous works \cite{langbridge2024optimal, calmon2017optimized} overcome this by learning a repair on a small training data set which can be applied to large archival or online data, analogous to generalisation in machine learning.

However, these repairs are highly sensitive to representation biases in the training data where the generalisation performance on under-represented classes is poorer due to incomplete learning of the distribution of the underlying data generating process. Representation bias is common in fairness settings, since disadvantaged classes of individuals have historically been denied access to opportunities. For example, in the Adult Income dataset \cite{adult}, non-white people have on average a lower level of education than the white people in the dataset, which corresponds to lower predicted salaries. The presence of representation bias in data can inherently limit the fairness metrics which are attainable, even after repair \cite{kleinberg2016inherent}. Additionally, the presence of multiple behaviour modes of the data-generating process -- due to intersectionality \cite{foulds2020intersectional} -- and the response of further segmenting training data (i.e. considering the outcomes of \textit{non-white women} rather than just women in the Adult Income example above) exacerbates these issues with statistical power, repair validity and learning \cite{ghosh2021characterizing}.

In this work, we propose a data-driven method for overcoming representation bias in fairness repair, extending the repair method initially proposed in \cite{langbridge2024optimal} to improve robustness to unbalanced data and improve the generalisation of the method. 

Our notational conventions are as follows: random variables (rvs) and their realizations,  e.g.\ $x$, are not distinguished notationally; column vectors, e.g.\  $\mathbf{x}$, are bold-faced; sets are denoted via blackboard symbols, $\mathbb{R}, \mathbb{N}$, etc.;  and functions via  sans-serif letters, $\O{T}(\cdot)$, $\O{Q}(\cdot)$, etc.  $\O{F}$  is reserved for an unspecified (i.e.\ wildcard) probability measure with respect to a particular algebra, and may also be used to refer to  its density, $\O{F}(\cdot)$, or probability mass function, $\Pr[\cdot ]$, with respect to the appropriate reference measure (Lebesgue or counting, respectively), as will be evident from the context. $\hat{\O{F}}_n$ denotes the empirical estimate of $\O{F}$ based on a random sample of size $n\geq 1$. Finally, the Dirichlet nonparametric process prior is denoted by $\mathcal D$. 

The paper is laid out as follows: in Section 2, we focus on modelling and introduce our (un)fairness criterion. Sections 3 and 4 detail our proposal for a data-driven fairness correction scheme, with experimental evidence supporting our method in Section 5. Section 6 contains a discussion of these results, and we conclude in Section 7.



\section{AI Unfairness and Representation Bias}\label{sec:modelling}
Consider a sequential observational experiment, in which continuous-valued, $d$-dimensional data (feature vectors), $x_i \in {\mathbb R}^d$, $i\in \mathbb{N}$, are made available, each labelled with Bernoulli (i.e.\ binary, without loss of generality) attributes, $u_i \in \{1,0\}$ (an {\em unprotected\/} attribute) and $s_i \in \{1,0\}$ (a {\em protected\/}, i.e.\ sensitive attribute). Examples of these kinds of attributes are provided in Section~\ref{sec:adult}. In this paper, we will gather a fully labelled  {\em research data set}, ${\mathbf z}_n$, of $n$ data. These fulfil the role of the training data in machine learning. They comprise $n_{u,s}$ data for each respective $(u,s)$-subgroup$^1$ \footnote{$^1$In this paper, the segmentation of $\mathbf z$ (\ref{eq:sntotal}) by $u$ will yield $u$-indexed {\em groups}, ${\mathbf z}_u$, each of which is further segmented into $(u,s)$-indexed {\em subgroups}, ${\mathbf z}_{u,s}$.}  of data, ${\mathbf z}_{u,s}$; i.e.:
\begin{eqnarray}
z_i &\equiv& \{x_i,u_i,s_i\}, \;\;i\in \{1, \ldots, n\}\nonumber\\
{\mathbf z}_{u,s} &\equiv& \{ z_i : (u_i,s_i)=(u,s)\}\;\; \#{\mathbf z}_{u,s} \equiv n_{u,s}\label{eq:usbatch}\\
{\mathbf z} &\equiv& \{{\mathbf z}_{u,s} : (u,s) \in \{1,0\}\times \{1,0\}\} \nonumber\\
n&\equiv& \sum_{u,s} n_{u,s}. \label{eq:sntotal}
\end{eqnarray}
The $n_{u,s} \geq 1$ are called the stopping numbers and will be learned from the data (Section~\ref{sec:stopping}). For notational ease, we may suppress the indexing variable, so that $z \equiv z_i$.  When convenient, we will also denote the labelled data, $z$, via $x_{u,s}$.  

The  causal  probability model  \cite{langbridge2024optimal} relating the attributes (i.e.\ labels) to the features (i.e.\ data) is (Figure~\ref{fig:ai-unfair}) 
 \begin{equation}
    \O{F}(x,u,s) \equiv \O{F}(x \vert u, s) \overbrace{\Pr[u \vert s] \Pr[s]}^{p_{u,s}}.
\label{eq:prob_model}
\end{equation}
We enforce the usual machine learning assumption of static labelled data, i.e.\ the feature vector field, $\mathbf{x}$, is conditionally independent and identically distributed (ciid) given the observed labels, ${\mathbf u}$ and ${\mathbf s}$:
\begin{equation}
    \O{F}({\mathbf x} | {\mathbf u}, {\mathbf s}) \equiv \prod_{i=1}^n \O{F}(x_i | u_i, s_i). \label{eq:static}
\end{equation}
Here, the $\O{F}_{u,s} \equiv \O{F}(x \vert u, s)$ are the $u,s$-class-conditional observation models (pdfs or pmfs) for the  features (continuous or discrete),  which must be learnt (generally nonparametrically) via the respective $(u,s)$-segmented research data (sub)sets, ${\mathbf z}_{u,s}$. As we accumulate the labelled observations, $z_i$, we will quench (i.e.\ stop) learning each of these models only when we satisfy a nonparametric stopping criterion (Section~\ref{sec:stopping}) for each component, reaching the data-driven stopping numbers,  $n_{u,s}$ (\ref{eq:usbatch}), respectively. 

\subsection{Societal and AI Unfairness}
\label{sec:unfair}

 \begin{figure}[ht]
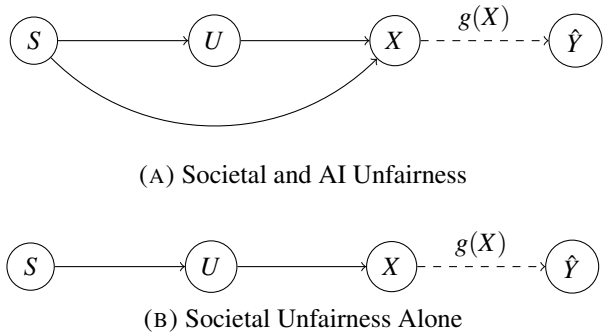

    \centering
    \vspace{-1.5em}
    \subfloat[Societal and AI Unfairness\label{fig:ai-unfair}]{
        \tikz [scale=1.2] {
        \node (s) [circle, fill=white, draw] at (0,0) {$S$};
        \node (u) [circle, fill=white, draw] at (2,0) {$U$};
        \node (x) [circle, fill=white, draw] at (4,0) {$X$};
        \node (y) [circle, fill=white, draw] at (6,0) {$\hat{Y}$};
        \graph { (s) -> (u) -> (x) ->[dashed, edge label=$g(X)$] (y), (s) ->[bend right=45] (x) };
        }}\\
    \subfloat[Societal Unfairness Alone\label{fig:ai-fair}]{\tikz [scale=1.2] {
        \node (s) [circle, fill=white, draw] at (0,0) {$S$};
        \node (u) [circle, fill=white, draw] at (2,0) {$U$};
        \node (x) [circle, fill=white, draw] at (4,0) {$X$};
        \node (y) [circle, fill=white, draw] at (6,0) {$\hat{Y}$};
        \graph { (s) -> (u) -> (x) ->[dashed, edge label=$g(X)$] (y) };
        }}
    \vspace{-0.5em}
    \caption{Comparison of the causal graphs of different sources of unfairness under our conditional independence model.}
    \label{fig:graphmodels}
\end{figure}

Two possible models for the static labelled observations, $\O{F}(z_i) \equiv \O{F}(z)$, are illustrated in Figure~\ref{fig:graphmodels}. Figure \ref{fig:ai-unfair} demonstrates the presence of AI unfairness, driven by the direct edge (i.e.\ learning pathway) between $s$ and $x$, and societal unfairness, driven by the edge between $s$ and $u$. The details of these definitions are provided in \cite{langbridge2024optimal}. In order to alleviate AI unfairness, we must break the link between $s$ and $x$, enforcing independence of the two variables \textit{conditional on $u$}. In Figure \ref{fig:ai-fair}, the conditional independence criterion $(x \independent s) \vert u$ is satisfied.

This conditional definition of fairness, introduced in \cite{kamiran2013quantifying} and extended in our previous work \cite{langbridge2024optimal}, distinguishes between discrimination which is \textit{illegal} (i.e. based on the sensitive attribute $S$), and that which is \textit{explainable} (i.e. driven by some other influential variable which is crucially not protected). This conditional definition is less stringent than unconditional independence, which means that data needs to be repaired less than under an unconditional scheme, while still satisfying legal definitions of fairness \cite{koumeri2023compatibility}.


In common with \cite{langbridge2024optimal}, the principal aim of this paper is statically to transform the labelled research data, $z \equiv x_{u,s}$ (\ref{eq:usbatch}), via $(u,s)$-indexed (surjective) transformations, $\O{T}_{u,s}$, in order to switch off these unfairness signals in the data, i.e.\ these lingering dependences of $x$ on $s$, given each state of $u$ (Fig.~\ref{fig:graphmodels}(a)). Under the  ciid and labelled assumptions, these transformations can then be applied to remove AI unfairness from out-of-sample data (e.g.\ archival data from the same generative process, and, therefore, manifesting AI unfairness). In summary:
\begin{equation}
x_{u,s} \stackrel{\O{T}_{u,s}}{\longrightarrow} x'_{u,s}
\equiv x'_u \;\;\O{s.th.} \;\;x'_u \sim \O{F}( x' \vert u,s) \equiv \O{F}(x' \vert u).
\label{eq:Tu}
\end{equation}
Here---by construction of the $(u,s)$-indexed transformations, $\O{T}_{u,s}$---the image, $x'_u$, is conditionally independent of its sensitive attribute, $s$, given its unprotected attribute, $u$. 
In Section~\ref{sec:fairness}, we will use methods of optimal transport (OT) to design the $\O{T}_{u,s}$, $u\in\{1,0\}$.

\subsection{Representation Bias}
Since the class-conditional distributions, $\O{F}_{u,s}$, are unknown (being nonparametric processes in the general case), they must first be inferred from the respective training data sets (i.e.\ subgroups), ${\mathbf z}_{u,s}$ (\ref{eq:usbatch}), typically via the empirical estimates,
\begin{equation}
\hat{\O{F}}_{\delta, n_{u,s}} \equiv \frac{1}{n_{u,s}} \sum_i \delta_{x_{u,s,i}},
    \label{eq:emp}
\end{equation}
where $\delta_x$ denotes the Dirac or Kronecker measure at $x$, as appropriate. 
In many AI fairness (AIF) settings, $u$ and $s$ are multivariate, i.e.\ many attributes are posited. This, of course, induces an exponentially increasing number of $(u,s)$-indexed subgroups (a manifestation of the curse-of-dimensionality,  known as {\em intersectionality\/} in the AIF literature \cite{foulds2020intersectional}). Over-segmentation of the $n$ research data may yield $u,s$-indexed batches which are  too small to ensure that learning of all the components, $\O{F}_{u,s}$, is complete, an issue we call {\em dilution}. Even if intersectionality is avoided (as in the current paper, where $u$ and $s$ are assumed to be {\em scalar\/} Bernoulli r.v.s), dilution remains an issue if any of the (predictive) subgroup probabilities, $p_{u,s} \equiv \Pr[u,s]$ (\ref{eq:prob_model}), are inherently small ($\ll \frac{1}{4}$). A central concern of this paper is to decouple the $n_{u,s}$ from the $p_{u,s}$ by implementing a Bayesian sequential decision-making scheme (i.e.\ stopping rule, Section~\ref{sec:stopping}) which ensures that research data continue to be accumulated until learning of all $\O{F}_{u,s}$ is complete  (in the sense defined in Section~\ref{sec:stopping}). 

\begin{definition}[Representation bias]
The $(u,s)$-subgroup suffers representation bias in a  set of $n$ ciid data if
\[ 
p_{u,s} n < n_{u,s},
\]
where $n_{u,s}$ is the stopping number for learning $x_i \stackrel{\O{ciid}}{\sim} \O{F}_{u,s}$.
\hfill $\square$
\label{def:urb}
\end{definition}

Many current methods for quantifying representation bias rely on approximately equalizing the representation of subgroups (i.e. enforcing$^2$ 
\footnote{$^2$We adopt the common convention of assigning $s=1$ to the } 
$\frac{\Pr[s=0 \vert u]}{\Pr[s=1 \vert u]} \geq \tau_u \;\forall \;u \in \{ 1,0\}$, where $\tau_u \in (0,1)$ is some threshold (lower bound) of \textit{allowable bias} in the representation of the $s=0$ subgroup in the $u$th group \cite{celis2020data}). Alternative definitions rely on the notion of \textit{coverage}, where there must be a minimum number of samples for each $(u,s)$-subgroup irrespective of the numbers for all other subgroups \cite{shahbazi2023representation}. A limitation of these approaches is that the thresholds must be set \textit{a priori}.

Via Definition~\ref{def:urb}, the research data size, $n\equiv \sum_{u,s} n_{u,s}$ (\ref{eq:sntotal}), ensures that
\begin{itemize}
    \item the subgroup sizes, $n_{u,s}$, are a (stochastic) function of both the class probabilities, $p_{u,s}$, {\em and\/} the class models, $\O{F}_{u,s}$ (\ref{eq:prob_model});
    \item learning of each $\O{F}_{u,s}$ is completed, in the sense of the stopping rule to be explained in Section~\ref{sec:stopping}; i.e.\  representation bias is avoided in {\em every\/} $(u,s)$-subgroup, including those with intrinsically low $p_{u,s}$.
    \end{itemize}
The proposed stopping rule will achieve {\em minimally sufficient} representation of each subgroup (as will be seen in Equation~\ref{eq:SR}). This satisfies another objective: to minimize the computational complexity of designing and applying the OT-based repairs, $\O{T}_{u,s}$ (\ref{eq:Tu}). Indeed, the availability of labelled research data---particularly those labelled by protected attributes, $s$---often requires expensive data-gathering procedures, legislative protection (such as that provided by the recent AI Act of the European Union \cite{AI_act}), and the adoption of arduous privacy-protecting measures.  
We now outline our scheme for sequential design of the stopping numbers, $n_{u,s}$.

\section{Bayesian Learning of the Sub-Group Models, $\O{F}_{u,s}$ }\label{sec:stopping}

For the present, consider any one of the (four) interleaved $u,s$-indexed learning processes, in each of which we sequentially observe $x_k \stackrel{\O{ciid}}{\sim} \O{F}(x_k | u,s) \equiv \O{F}_{u,s}$, being the (four) components of the mixture model in (\ref{eq:prob_model}). For notational convenience in this section,  we suppress the $u,s$ subscripts, and have used $k \in\{1,2,\ldots\}$ to index the sub-sequence of $u,s$-indexed features. We also assume that $x_k \in \mathbb{R}$, i.e.\ that $d=1$ and that the feature is continuous. All of these assumpations can be readily relaxed if required. 

Since $\O{F}$ is an unknown distribution (here, a pdf), we model it as a nonparametric process, equipped with an appropriate Bayesian nonparametric (BNP) process prior, $\mathcal F$~\cite{Ghos:17}, yielding the hierarchy, 
\begin{equation}
  x\sim \O{F}\;\; \O{and} \;\; \O{F}\sim {\mathcal F}. \label{eq:BNPhier}  
\end{equation}
 Specifically, we adopt the Dirichlet nonparametric process prior (DPP) \cite{ferguson1973bayesian}, $\O{F} \equiv {\mathcal D}(\hat{\O{F}}_0, \nu_0)$, where $\hat{\O{F}}_0 \equiv \O{E}_{\mathcal D}[\O{F}]$ is the prior expected distribution of $x$ and $\nu_0 \in {\mathbb R}^+$ is the prior degrees-of-freedom (d.o.f.) parameter. 
Both of these parameters must be specified {\em a priori}. Since our uncertainty about $\O{F}$ is a monotonically decreasing function of $\nu_0$ \cite{ferguson1973bayesian}, a minimally informative choice is a diffuse $\hat{\O{F}}_0$ (such as a quasi-uniform distribution on $\mathbb R$, and $\nu_0 \downarrow 0^+$ \cite{quinn2007learning}.

The appropriate BNP stopping rule for learning the mixture components, $\O{F}$, is provided in \cite{quinn2007learning}. It emerges via two key properties of the DPP, $\mathcal D$~\cite{ferguson1973bayesian}:
\begin{itemize}
    
    \item[(i)] $\mathcal F$ is the conjugate distribution \cite{BernSmi:10} for learning $\O{F}$ via ciid sampling:
    \begin{eqnarray}
    \O{F} | \{x_i, \ldots, x_k\} &\sim& {\mathcal D}(\hat{F}_k, \nu_k)\equiv {\mathcal D}_k,\label{eq:DPP}\\
    \hat{F}_k &=& (\nu_0+k)^{-1}\left(\nu_0\hat{F}_0 + \sum_{j=1}^k \delta(x_j)\right),\nonumber\\
    \nu_k &=& \nu_0+k,\nonumber
    \end{eqnarray}
    confirming the concentration of $\mathcal D$ around its mean distribution, $\hat{F}_k$, as $k\uparrow \infty$. Note that $\hat{F}_k$ is the Bayesian generalization of the classical empirical estimate of $\O{F}$, 
    \begin{eqnarray}
\hat{F}_{\delta, k} &\equiv&  \frac{1}{k}\sum_{j=1}^k \delta(x_j), \nonumber\\
\alpha_k &\equiv& \frac{k}{\nu_0 +k},\nonumber\\
         \hat{F}_k &=& (1-\alpha_k) \hat{F}_0 + \alpha_k \hat{F}_{\delta, k}, \label{eq:expF}    
    \end{eqnarray}
    i.e.\ the $\alpha_k \in (0,1)$-weighted mixture of the prior estimate, $\hat{F}_0$, of the unknown distribution and the empirical estimate, $\hat{F}_{\delta, k}$, based on the sequential realizations, ${\mathbf x}_k$. 
 
 \item[(ii)] $\O{F}$ induces an unknown multinomial distribution, ${\mathbf p} \in \Delta$, on any finite, measurable partition of the codomain, $\mathbb R$, of $x_k \equiv x$. In the case where $\O{F}\sim {\mathcal D}$, then the distribution induced on the simplex, $\Delta$, is the Dirichlet distribution, ${\mathbf p} \in \O{D}$.      
\end{itemize}
The finite measurable partition in (ii) can be thought of as a {\em quantization}, $\O{Q}_{\mathbb V}$, of $x \stackrel{\O{ciid}}{\sim} \O{F}$:
\[
x \stackrel{\O{Q}_{\mathbb V}}{\longrightarrow} x'  \equiv \O{Q}_{\mathbb V}(x), 
\]
where $\mathbb V$ is the pre-defined set of vertices of the quantizer. Here---as in~\cite{quinn2007learning}---$\mathbb V$ is sequentially defined as the ordered observations, $x_{(j)}$, themselves:
\begin{eqnarray}
    {\mathbb V}_0 &\equiv& \{-\infty, +\infty\}\nonumber\\
    {\mathbb V}_k \equiv \{-\infty, \sigma_1, \ldots, \sigma_k, +\infty\} &\equiv& {\mathbb V}_{k-1} \cup \{x_k\}, \;\; k=1,2,\ldots,\nonumber\\
    \sigma_j &\equiv& x_{(j)}.\label{eq:PRS}
\end{eqnarray}
Note that the $x_k \stackrel{\O{ciid}}{\sim} \O{F}$ are a.s. distinct in the assumed case where  $\O{F}$ is a continuous measure. In the discrete case, ${\mathbb V}_k $ assembles all distinct observed states of $x$ (i.e.\ no repetitions), and so $\# {\mathbb V}_k -2 \leq k$ in general.  

This sequential, data-driven (re)quantization process, ${\mathbb V}_k$, $k=1,2,\ldots$, constitutes a {\em partition refinement schedule}, in the sense that the number,  $\# {\mathbb V}_k -2$, of ordered vertices, $\sigma_j$, is an increasing function of the number, $k$, of ciid observations, ${\mathbf x}_k \equiv \{x_1, \ldots, x_k\}$. As $k\uparrow\infty$, knowledge of the underlying (typically continuous) unknown distribution, $\O{F}\sim {\mathcal D}_k$ (\ref{eq:DPP}), concentrates around its sequential expected value, $\hat{\O{F}}_k$ (i.e.\ the prior-empirical weighted mixture (\ref{eq:expF})). Learning can be quenched under a suitable stopping rule (criterion). In~\cite{quinn2007learning}, this is constructed via the sequence of   Kullback-Leibler divergences (KLDs)~\cite{KLD51},  $\O{KLD}[\O{D}_k || \O{D}_{k-1}]$,   $ k\geq 2$,  i.e.\ via the sequence of Dirichlet distributions, $\O{D}_ k$    (such that ${\mathbf p}_k \sim \O{D}_k$ ), induced via the proposed data-driven partition refinement schedule (\ref{eq:PRS})$^3$. \footnote{$^3$The (a.s.\ in the continuous case) sequential splitting of mass in the multinomial process, ${\mathbf p} _k$, is a type of stick-breaking process~\cite{Seth:94} for representation of the underlying DPP (\ref{eq:DPP}). Importantly, the splitting probabilities are data-driven here, being controlled by the relationship between $x_k$ and ${\mathbf x}_{k-1}$.} The stopping rule is therefore defined via a threshold, $\epsilon >0$:
\begin{equation}
\hat{n} \equiv \min_{k\in{\mathbb N}^+} \{k : \O{KLD}[\O{D}_k || \O{D}_{k-1}] < \epsilon\}. \label{eq:SR}
\end{equation}
In practice, a smoothed version of the KLD sequence (\ref{eq:SR}) is adopted. The details are provided in~\cite{quinn2007learning}, along with the implied algorithm. 

Recall that this BNP learning and stopping procedure is applied, in parallel, to all four components, $\O{F}_{u,s}$, of the attribute-structured mixture model  (\ref{eq:prob_model}), yielding attribute-dependent stopping numbers, $\hat{n}_{u,s}$, $(u,s)\in \{1,0\}\times\{1,0\}$. The stopping thresholds can be chosen attribute-wise, i.e.\ $\epsilon_{u,s}$. These provide a set of operating conditions which trade off sample complexity against accuracy of the BNP learning process for the $\O{F}_{u,s}$ (\ref{eq:BNPhier}).   

\section{Data-Driven Fairness Correction}\label{sec:fairness}
In this section, we propose an adaptation of the distributional approach for data repair in \cite{langbridge2024optimal} to  overcome representation bias using the nonparametrically learnt subgroup models, $\O{F}_{u,s}$ (Section \ref{sec:stopping}). This is a fundamentally different paradigm from other approaches to data repair \cite{feldman2015certifying, gordaliza2019obtaining, dwork2012fairness}, which do not comment on the selection of training data and hence representation bias may be amplified through the repair process.

\subsection{Learning the Repair Operation}
More formally, our labelled and ciid research (i.e.\ training) data set, $\mathbf{z}$, is defined according to the quenching (i.e.\ stopping) of learning (\ref{eq:SR}) in each $(u,s)$-subgroup (\ref{eq:sntotal}). We consider the set of centroids of each interior cell, i.e. the arithmetic means of the interior vertices $\sigma \in \mathbb{V}_{u,s}$:
\begin{equation}
    q_{u,s,j} \equiv \frac{\left( \sigma_j + \sigma_{j+1} \right)}{2}, 1 \leq j \leq \hat{n}_{u,s} - 1, \label{eq:q_u,s}
\end{equation}
where $q_{u,s,j} \in \mathbb{Q}_{u,s}$ is the centroid of the $j$\textsuperscript{th} cell. 

Under the proposed partition refinement schedule (\ref{eq:PRS}), the quantized conditionals,  $\mu_{u,s}$, are uniform on the ordered set $\mathbb{Q}_{u,s}$, such that
\begin{eqnarray}
    \mu_{u,s} &\equiv& \frac{1}{\hat{n}_{u,s} - 1} \sum_j \delta_{q_{u,s,j}}.\label{eq:mu_u,s}
\end{eqnarray}

Note that this approach is distinct from the kernel density approximation method in \cite{langbridge2024optimal}, since we consider a uniform $\mu_{u,s}$ over a support defined by the observations $\mathbf{x}_{u,s}$ (\ref{eq:q_u,s}). Following \cite{langbridge2024optimal}, we employ techniques from optimal transport (OT) to design our repair. For a detailed treatment of OT theory, and further background related to OT for fairness, see \cite{peyr:19} and \cite{feldman2015certifying, chzhen2020fair, gordaliza2019obtaining} respectively.



The Wasserstein distance from $\mu_{u,0}$ to $\mu_{u,1}$ with respect to the cost function $\mathsf{C}(q_{j_0}, q_{j_1}) \equiv \O{L}_p^p$, the latter being the $p$\textsuperscript{th} power of the $\O{L}^p$ norm (with $p\equiv 2$ in this work):
\begin{equation*}
    \O{W}_p^p \left( \mu_{u,0}, \mu_{u,1} \right) \equiv \min_{\pi \in \Pi \left( \mu_{u,0}, \mu_{u,1} \right) } \sum_{q_{j_1}} \sum_{q_{j_0}} \mathsf{C} \left(q_{j_0}, q_{j_1} \right) \pi \left(q_{j_0},q_{j_1} \right).
\end{equation*}
Here, $\Pi (\cdot)$ is the set of joint pmfs  with marginals $\mu_{u,0}$ and $\mu_{u,1}$, defined on the product space, $\mathbb{Q}_{u,0} \times \mathbb{Q}_{u,1}$. 

The $t$-indexed Wasserstein barycentres of $\mu_{u,0}$ and $\mu_{u,1}$ are
\begin{equation}
 \label{eq:bary}
 \nu_{t,u} \equiv \min_{\nu} \left\{ (1 - t)\O{W}_2^2(\mu_{u,0}, \nu) + t\O{W}_2^2(\mu_{u,1},\nu) \right\}, t \in [0, 1],
\end{equation}
defining the geodesic between the two empirical marginals.

We consider the centre of the geodesic, $\nu_{u} \equiv \nu_{0.5,u}$, as our fair $u$-conditional target design, since it is $s$-invariant by definition, and equally incurs damage---in a sense to be defined in Section~\ref{sec:damage}---to both $s$-subgroups simultaneously. We compute $\nu_u$ via the $(\hat{n}_{u,0} - 1) \times (\hat{n}_{u,1} -1)$ OT plan between the $(u,s)$-conditional quantized distributions, $\mu_{u,s}$:
\begin{equation}
    \pi^*_u \equiv \argmin_{\pi \in \Pi \left( \mu_{u,0}, \mu_{u,1} \right) } \sum_{q_{j_1}} \sum_{q_{j_0}} \mathsf{C} \left(q_{j_0}, q_{j_1} \right) \pi \left(q_{j_0},q_{j_1} \right).
    \label{eq:kan_plan}
\end{equation}


Now, consider a labelled draw $x_{u,s} \stackrel{\O{ciid}}{\sim} \O{F}_{u,s}$. Our task  is to repair this datum by driving it to the corresponding point $x' \sim \nu_u$ on the barycentre (\ref{eq:Tu}). To achieve this, we define a stochastic operator $\O{T}_{u,s}$ with two sources of randomness as follows:
\begin{itemize}
    \item[(i)] Denoting the round-down (i.e.\ truncation) state of $x$ in $\mathbb{Q}_{u,s}$ by $q_{j_s} \equiv \lfloor x \rfloor$, we design a Bernoulli trial $b \sim \mathcal{B}(p)$ where $p \equiv \frac{x - q_{j_s}}{q_{j_s + 1} - q_{j_s}}$. Then,  $j_s \leftarrow j_s + b$  defines the appropriate row (where $s = 0$) or column (where $s = 1$) of the transport plan $\pi^*_u$.
    \item[(ii)] Let $\pi^*_{u;j_0,:}$ be the $j_0$\textsuperscript{th} row (or $\pi^*_{u;:,j_1}$ the $j_1$\textsuperscript{th} column) of $\pi^*_u$. Depending on the value of $s\in\{1,0\}$, this row (or column) defines the probability vector of a multinomial trial. For $s=0$, $\Pr[j_1 | u, s=0, x] \propto \pi^*_{u;j_0, j_1}$, yielding the (randomly) repaired state, $q_{j_1}$. Correspondingly, for  $s=1$,  $j_1$ provides the column index into $\pi^*_u$, yielding the multinomial for  realizing the (randomly) repaired state, $q_{j_0}$.
\end{itemize}
Finally, the OT repair (\ref{eq:Tu}), mapping to the geodesic centre ($t=0.5$) (\ref{eq:bary}), is
\begin{equation}
    x_{u,s} \rightarrow x'_u \equiv \O{T}_{u,s}(x_{u,s}) \equiv 0.5 q_{j_0} + 0.5 q_{j_1}
\label{eq:repair}
\end{equation}




\subsection{Evaluating (Un)Fairness} Clearly, there are different degrees of fairness (Section~\ref{sec:unfair}) associated with the pre- ($\mathbf{x}$) and post-repair ($\mathbf{x'}$) data. We would like to quantify this fairness. To this end, we adapt the KLD-based fairness metric proposed by \cite{langbridge2024optimal}. Unlike other widely-adopted fairness metrics \cite{narayanan21fairness, friedler2019comparative, zafar2017fairness}, this measure does not depend on classification outcome (and therefore is model-invariant), rather it is a function of the complete distribution (Equation~\ref{eq:prob_model}).

As proposed in \cite{langbridge2024optimal}, we evaluate the $s$-dependence of the $u$-conditional distributions using the $u$-expectation of the symmetrized Kullback–Leibler Divergence (KLD), $E(\mathbf{x})$: 
\begin{eqnarray}
E_u(\mathbf{x}) &\equiv& \frac{1}{2} \O{D} \biggl[ \O{F}_{u,0} \  \big|\big| \  \O{F}_{u,1} \biggr] + \frac{1}{2} \O{D} \biggl[ \O{F}_{u,1} \  \big|\big| \  \O{F}_{u,0} \biggr], \nonumber\\
E(\mathbf{x}) &=& \sum_u \Pr[u] \  E_{u}(\mathbf{x}). \label{eq:E-u}
\end{eqnarray}
The smaller this quantity, the fairer the distribution in the sense of breaking the conditional dependence between $x$ and $s$, given $u$ (Figure~\ref{fig:graphmodels}). 
We calibrate the  unfairness after repair, $E(\mathbf{x}')$, against the unfairness of the original data, yielding the summary metric, $\hat{E}$:
\begin{equation}
    \hat{E} \equiv \frac{E(\mathbf{x}')}{E(\mathbf{x})}
\label{eq:E-hat}
\end{equation}
For a competent repair, $\hat{E}$ should be less than 1, with $\hat{E} = 1$ corresponding to no improvement over the original data, and $\hat{E} = 0$ indicating total $s$-invariance (fairness). 

\subsection{Data Damage}
\label{sec:damage}
Unfair data are being transformed in order to satisfy notions of fairness. A trivial way to ensure fairness according to (\ref{eq:E-u}) would be to enforce $x'_{u,s} \equiv \O{const.}  \; \forall (u,s) \in \{1,0\} \times \{1,0\}$, 
However, this destroys all predictive information in the features. We introduce a novel metric to evaluate this notion of \textit{data damage}.

Following principles of information projection \cite{cover1999elements} and fully probabilistic design (FPD) \cite{karny2012axiomatisation}, we evaluate the KLD from the repaired (fair) data distribution to the original (unfair) data distribution, in order to measure the degree to which data have been damaged by the OT repair operation (\ref{eq:repair}).
\begin{definition}
    \begin{equation}
    \label{eq:usdamage}
    D_{u,s} \equiv \O{D} \biggl[ \O{F}_{u,s} \  \big|\big| \  \O{F}'_{u,s}  \biggr],
    \end{equation}
\label{def:kld_damage}
where $\O{F}'_{u,s} \equiv \O{F}(x' | u,s)$.
\end{definition}
A summary quantifier of the damage incurred in the repaired data distribution, $\O{F}(x')$,   the $(u,s)$-expected value of 
$D_{u,s}$ is evaluated:
\begin{equation}
    D = \sum_u \Pr[u] \sum_s \Pr[s | u] \ D_{u,s}.
\label{eq:kld-u}
\end{equation}
Once again, smaller is better:  $D \downarrow 0$ represents data which have been damaged less.

\section{Experiments}\label{sec:exp}
In this section, we detail a number of experiments designed to validate the key proposals of this work. In Section \ref{sec:stop-exp}, we explore the behaviour of the stopping rule proposed in Section \ref{sec:stopping} with quantized and mixture data in order to assess it in realistic fairness problems. Then, in Section \ref{sec:rb-exp}, we evaluate the performance of our repair method with respect to (\ref{eq:E-hat}) and (\ref{eq:kld-u}) under varying mixture probabilities, $Pr[u,s]$, to simulate the impact of representation bias on our repair. Section \ref{sec:benchmarking-exp} contains the results of benchmarking against two SOTA data repair approaches: geometric repair \cite{feldman2015certifying, gordaliza2019obtaining} and distributional repair \cite{langbridge2024optimal}.

Throughout the experiments, unless otherwise specified, we set the prior confidence $\nu_0 \equiv 0.001$, the stopping criterion $\varepsilon \equiv 0.01$, we take the uniform prior $\hat{F}_0 \equiv \mathcal{U}(x_\text{min}, x_\text{max})$, and we assume scalar data (features), i.e.\ $d\equiv 1$.

\pagebreak

\subsection{Learning non-Gaussian Models}\label{sec:stop-exp}

\subsubsection{Multinomial Models}
For quantized data, the probability of coincident observations is non-zero. To evaluate the performance of the stopping rule in this setting, we construct categorical distributions with $q$ categories, where the categorical probability is based on some reference measure $X$. We set $X \equiv \mathcal{N}(0,1^2)$, and define $p_i \equiv F_X(x_{i+1}) - F_X(x_i) \forall i \leq q$ where $F_X$ is the cumulative distribution function for $X$ and the support is defined $x \in \mathbb{Q}, \mathbb{Q} \equiv\{ -5.0, \cdots, -5.0 + \frac{10i}{q}, \cdots, 5.0\} , |\mathbb{Q}| \equiv q + 1$.

We evaluate the evolution of the log-KLD (LKLD) until stopping at $n \equiv \hat{n}$ (\ref{eq:SR}) for five categorical models with $q \in \{ 5, 10, 50, 500, 5000\}$. As is evident from Figure \ref{fig:multi_stop}, learning is quenched sooner for coarsely quantized data with low $q$. This suggests the stopping rule is able to exploit the information encoded in the coincident nature of samples to stop sooner.

 \begin{figure}[ht]
    \centering
    \includegraphics[width=0.65\columnwidth]{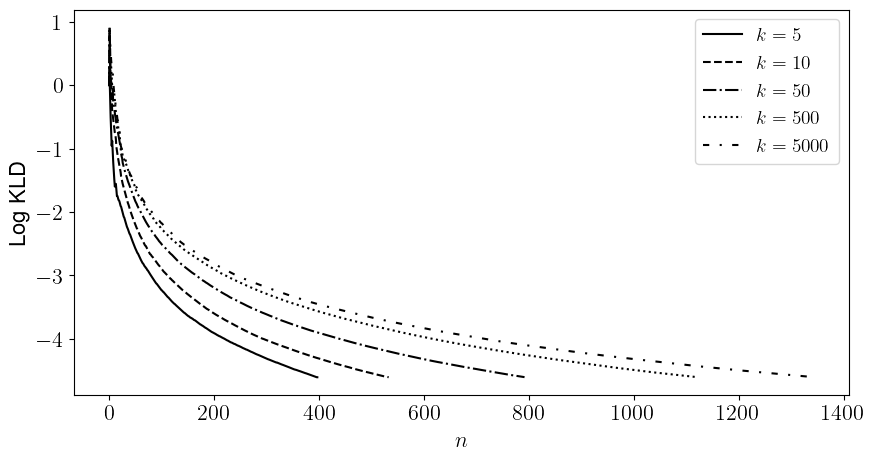}
    \vspace{-1em}
    \caption{LKLD until stopping for observations from categorical variables with $q$ states.} 
    \label{fig:multi_stop}
\end{figure}

\subsubsection{Gaussian Mixture Models}
We further validate the stopping rule's convergence on Gaussian mixture model (GMM) data with a minority component, where $\O{F} \equiv \sum_{i=1}^2 \alpha_i \mathcal{N}(m_i, \sigma_i^2)$ and $\mathbf{\alpha} \equiv [0.8, 0.2]$, ${\mathbf m} \equiv [-1, -5]$, ${\mathbf \sigma} \equiv [1, 0.5]$. This model simulates the presence of intersectionality \cite{foulds2020intersectional}, such that within $(u,s)$-conditional subgroups a mixture of components remains. In this section, we conduct three experiments: (i) we evaluate the effect of the prior confidence $\nu_0$ on stopping, (ii) the effect of a non-uniform prior, and (iii) we validate the behaviour of the stopping rule over 500 Monte-Carlo simulations.

Figure \ref{fig:nu_0} demonstrates that as $\nu_0 \rightarrow 1$, stopping occurs sooner since the prior has more influence over stopping. Where we are uncertain about the prior, a low $\nu_0 \approx 0.001$ ensures that stopping is data-driven.

The theoretical results in Section \ref{sec:stopping} are validated by our empirical findings in Figure \ref{fig:prior-dep}. The uniform prior is conservative (in the sense that it is minimally informative), with the Gaussian prior $\hat{F}_0 \equiv \mathcal{N}(m, 1^2)$ yielding higher stopping numbers, $\hat{n}$, when $x \not\sim \hat{F}_0$, i.e.\ when there is mismatch between the prior and the data generating process $\O{F}$ (\ref{eq:prob_model}).

 \begin{figure}[ht]
    \centering
    \vspace{-1em}
    \subfloat[Stopping numbers, $\hat{n}$, for varying prior weights, $\nu_0$ (and $\hat{\O{F}}_0$ uniform).\label{fig:nu_0}]{\includegraphics[width=0.39\columnwidth]{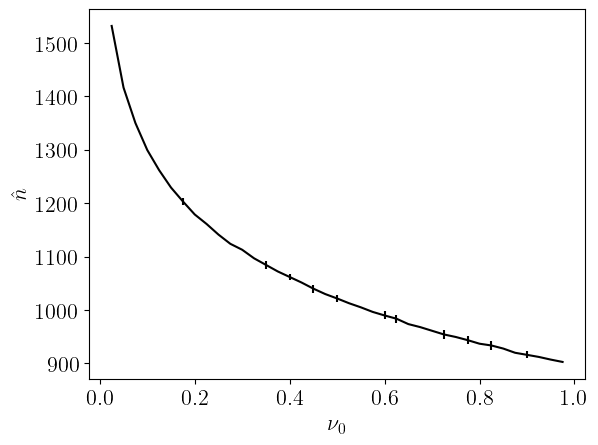}} \quad
    \subfloat[Stopping numbers, $\hat{n}$, for different prior means, $\hat{F}_0 \equiv \mathcal{N}(m, 1^2)$, compared to the uniform prior.\label{fig:prior-dep}]{\includegraphics[width=0.55\columnwidth]{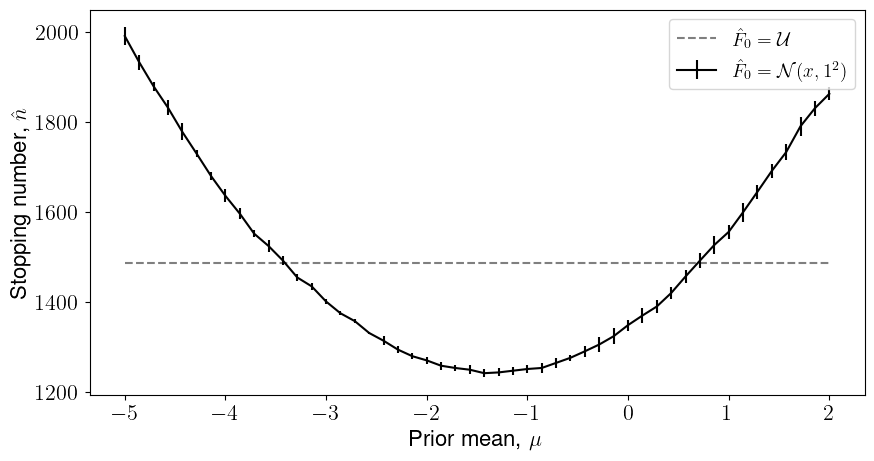}}
    \vspace{-0.5em}
    \caption{} 
\end{figure}

Figure \ref{fig:sample_stats} shows that the stopping numbers, $\hat{n}$, are independent of the individual random data realisations, with the stopping rule converging to the underlying mean and variance when learning is quenched. This is consistent with the results reported in \cite{quinn2007learning} for Student's t-distribution and supports the application of the stopping rule to the mixture-type data typical in machine learning applications. 

\begin{figure}[ht]
    \centering
    \includegraphics[width=0.9\columnwidth]{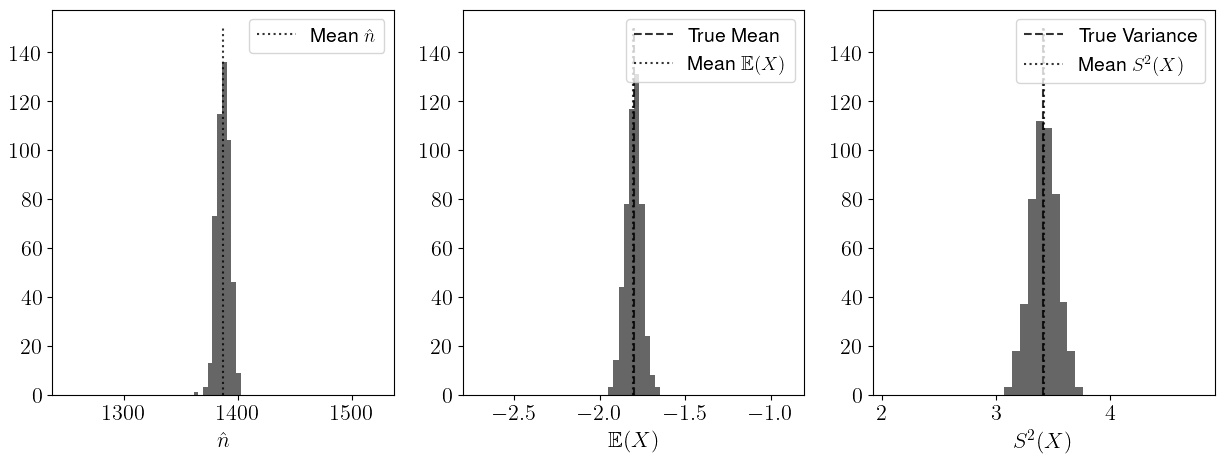}
    \vspace{-1em}
    \caption{Stopping numbers, $\hat{n}$, sample mean, $\mathbb{E}(X)$, and sample variance, $S^2(X)$, for GMM data over 500 Monte-Carlo simulations.}
    \label{fig:sample_stats}
\end{figure}

\subsection{Fairness Repair Under Representation Bias}\label{sec:rb-exp}
To evaluate the method's performance with varying levels of representation bias, we consider a setting with Gaussian components $F(x,u,s)$ as given in Table \ref{tab:rb_model}. The mixture weights, $p_{u,s}$, are chosen subject to the constraint that $\Pr[s|u] \equiv 0.5$, $u\in\{1,0\}$. The $u$-weight, $\Pr[U = 0]$, is then varied in $ (0, 0.5]$ to simulate balanced and unbalanced $u$ classes.

\begin{table}[ht]
    \centering
    \begin{tabular}{ll|l}
        \toprule
        $u$ & $s$ & $\O{F}(x|u,s)$\\
        \midrule
        0 & 0 & $\mathcal{N}(-1.0, 1^2)$ \\
        0 & 1 & $\mathcal{N}(1.0, 1.2^2)$ \\
        1 & 0 & $\mathcal{N}(-0.5, 1.2^2)$ \\
        1 & 1 & $\mathcal{N}(1.5, 0.8^2)$ \\
        \bottomrule
    \end{tabular}
    \caption{Model components (\ref{eq:prob_model}) in the  representation bias simulation. $p_{u,s}$ is varied throughout the experiment.}
    \label{tab:rb_model}
\end{table}

Figure \ref{fig:rb} demonstrates a key strength of our approach: even with significant representation bias in the data-generating process, where $Pr[U = 0] \equiv 0.025$, our data-driven approach is able to repair data reliably.
Further, repair damage is also shown to be representation bias-invariant, with no more damage to under-represented components than when representational parity is achieved at $Pr[U = 0] = 0.5$. A modified approach without the stopping rule, where $n_{u,s} \equiv p_{u,s}\hat{n}$, is unable to achieve comparable levels of repair even for $Pr[U=0] \approx 0.5$ (being the case in which representation bias is minimal) due to incomplete learning of subgroup models for some $(u,s)$.

\begin{figure}[ht]
    \centering
    \subfloat[$S$-invariance.]{\includegraphics[width=0.4\columnwidth]{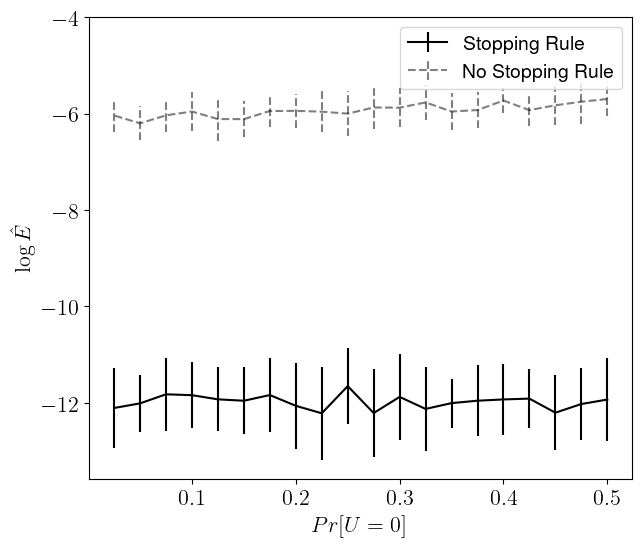}}
    \subfloat[Data damage.]{\includegraphics[width=0.4\columnwidth]{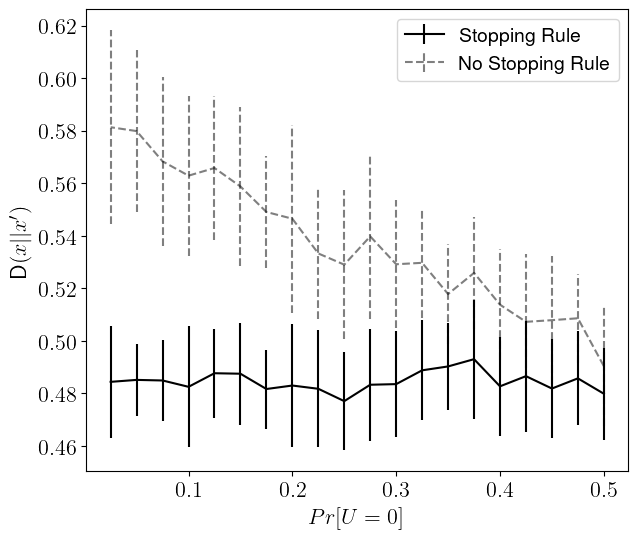}}
    \vspace{-1em}
    \caption{$\log \hat{E}$ and damage (Definition \ref{def:kld_damage}) for data with simulated representation bias $Pr[U = 0] \in (0, 0.5]$. Results are reported $\pm$ standard deviation over 20 Monte-Carlo simulations.}
    \label{fig:rb}
\end{figure}

\subsection{Benchmarking on Simulated and Real Data}\label{sec:benchmarking-exp}
In this section, we compare our method to two state-of-the-art approaches: the distributional repair proposed in \cite{langbridge2024optimal}, and the geometric repair proposed by \cite{feldman2015certifying} and extended by \cite{gordaliza2019obtaining}. Neither of these approaches deal explicitly with representation bias, and hence we evaluate these approaches on an alternative dataset, $\tilde{\mathbf{z}}_{u,s}$, defined as \ref{eq:sntotal} with $n_{u,s} \equiv p_{u,s} \hat{n}$ (i.e. retaining the representation bias in $u$ and $s$ of the overall model $\O{F}$ in a sample of equal size to the total stopping numbers).

\subsubsection{Simulated GMM Data with Intersectionality}
In order to simulate the impact of intersectionality and representation bias on different repair methods, we design a simulated mixture model dataset with $p_{u,s}$ imbalance parameterised as in Table \ref{tab:sim_fair_model}. Note the bias in both the mixing probabilities $p_{u,s}$ and the mixture weights in $\O{F}(x | s,u)$.

\begin{table}[ht]
    \centering
    \begin{tabular}{ll|ll}
        \toprule
        $u$ & $s$ & $p_{u,s}$ & $\O{F}(x | s,u)$\\
        \midrule
        0 & 0 & 0.18 & $0.8\mathcal{N}(-1.0, 1^2) + 0.2\mathcal{N}(-5.0, 0.5^2)$ \\
        0 & 1 & 0.12 & $0.6\mathcal{N}(1.0, 1.2^2) + 0.4\mathcal{N}(-1.75, 0.5^2)$ \\
        1 & 0 & 0.42 & $0.5\mathcal{N}(-1.0, 1^2) + 0.5\mathcal{N}(3.5, 1.2^2)$ \\
        1 & 1 & 0.28 & $0.1\mathcal{N}(-2.0, 0.8^2) + 0.9\mathcal{N}(5.0, 1.5^2)$ \\
        \bottomrule
    \end{tabular}
    \caption{Model Components for simulated intersectionality study.}
    \label{tab:sim_fair_model}
\end{table}

As Table \ref{tab:sim-results} shows, our approach is able to outperform both geometric \cite{feldman2015certifying, gordaliza2019obtaining} and distributional \cite{langbridge2024optimal} approaches with respect to the $s$-invariance metric $\hat{E}$ (\ref{eq:E-hat}) for both on- and off-sample repair. This is due to the complete learning of all $\O{F}_{u,s}$ components, ensuring that underrepresented $u,s$ subgroups are not neglected in the repair operation.

While distributional repair leads to the less damage than our approach, this is likely due to the higher remaining $s$-conditional dependence after distributional repair. The trade-off between the amount of repair and the predictive utility of data is explored in detail in \cite{gordaliza2019obtaining}.

\begin{table}[ht]
    \centering
    \begin{tabular}{l|cc|cc}
        \toprule
         & \multicolumn{2}{|c|}{On-Sample} & \multicolumn{2}{|c}{Off-Sample} \\
        Method & $\log \hat{E}$ & Damage & $\log \hat{E}$ & Damage\\
        \midrule
        Geometric \cite{feldman2015certifying, gordaliza2019obtaining} & $-8.365 \pm 0.172$ & $0.394 \pm 0.020$ & - & -\\
        Distributional \cite{langbridge2024optimal} & $-4.491 \pm 0.120$ & $\mathbf{0.307 \pm 0.016}$ & $-4.229 \pm 0.232$ & $\mathbf{0.325 \pm 0.016}$\\
        Our Approach & $\mathbf{-13.550 \pm 0.782}$ & $0.399 \pm 0.018$ & $\mathbf{-5.385 \pm 0.375}$ & $0.431 \pm 0.020$\\
        \bottomrule
    \end{tabular}
    \caption{Repair metrics $\hat{E}$ and $D$ for Geometric, Distributional and Stick-Breaking Repairs on a simulated GMM dataset over 500 Monte-Carlo trials. Note that $\hat{E}$ is reported on the log scale for clarity. Geometric repair cannot repair off-sample data.}
    \label{tab:sim-results}
\end{table}

\subsubsection{Adult Income Data}
\label{sec:adult}
We further validate our repair method on the Adult Income dataset \cite{adult}. The dataset consists of information about $n_0 \equiv 48,882$ individuals, including their age, sex, race, education level, and annual income. We consider sex as the protected attribute, with $S = 1$ corresponding to the males and $S = 0$ to the non-males. Our unprotected, explanatory attribute is education level, with $U = 0$ corresponding to an education level of high-school graduate or below. In this data, both representation bias (due to differing access to higher education between men and women) and intersectionality (due to the impact of race on education) are present. Each subgroup's prior probability is given in Table \ref{tab:adult_weights}.

\begin{table}[ht]
    \centering
    \begin{tabular}{ll|l}
        \toprule
        $u$ & $s$ & $Pr[u,s]$\\
        \midrule
        0 & 0 & $0.146$ \\
        0 & 1 & $0.310$ \\
        1 & 0 & $0.187$ \\
        1 & 1 & $0.357$ \\
        \bottomrule
    \end{tabular}
    \caption{Mixture weights for the Adult income data, stratified by sex and education level.}
    \label{tab:adult_weights}
\end{table}


Table \ref{tab:adult-results} summarises the $s$-conditional dependence remaining in the data after repair with (i) geometric repair and (ii) our method, and demonstrates that for on-sample repairs our method is able to match or outperform geometric repair, and is able to reduce the $s$-dependence present in unseen data by at least three times.

\begin{table}[ht]
    \centering
    \begin{tabular}{l|cc|cc}
        \toprule
         & \multicolumn{2}{|c|}{Geometric Repair} & \multicolumn{2}{|c}{Our Approach} \\
        Feature & On-Sample $\hat{E}$ & Off-Sample $\hat{E}$ & On-Sample $\hat{E}$ & Off-Sample $\hat{E}$\\
        \midrule
        Age & \textbf{0.0031} & - & 0.0041 & \textbf{0.1374}\\
        Capital Gain & 0.2222 & - & \textbf{0.1366} & \textbf{0.2103}\\
        Capital Loss & \textit{3.1058} & - & \textbf{0.5638} & \textbf{0.3018}\\
        \bottomrule
    \end{tabular}
    \caption{$\hat{E}$ for Geometric and Stick-Breaking Repairs on the Adult dataset. Note that Geometric repair cannot repair off-sample data, and that $\hat{E} > 1$ corresponds to making data \textit{less} fair with respect to $S$. 
    }
    \label{tab:adult-results}
\end{table}

\section{Discussion}
The results presented in Section \ref{sec:exp} demonstrate the performance of our approach on a number of diverse models, $\O{F}$. The experiments in Section \ref{sec:stop-exp} support the Bayesian non-parametric theory of Section \ref{sec:stopping} and results presented in \cite{quinn2007learning}, demonstrating that the stopping rule is appropriate for a number of data-generating processes, including discretised and mixture models. In this Section, we further validate that -- while the stopping rule is not prior-invariant -- the \textit{cost} of an inaccurate prior is not incomplete learning but rather a higher stopping number $\hat{n}$ (\ref{eq:SR}). The uniform prior, $\hat{F}_0 \equiv \mathcal{U}$, is shown to be a powerful choice when there is large uncertainty about $\O{F}$, such as the fairness setting when we have no knowledge about the data-generating process \textit{a priori}. Empirical guidance for the selection of the operating conditions of our method are given in Section \ref{sec:stop-exp}, and applied through Sections \ref{sec:rb-exp} and \ref{sec:benchmarking-exp}.

The key advantage of our approach under representation bias is demonstrated in Section \ref{sec:rb-exp}, where our data-driven stick-breaking approach is able to achieve representation-invariant repairs. Similarly, the results in Section \ref{sec:benchmarking-exp} demonstrate that our approach performs better than the SOTA on a simulated GMM dataset designed to mimic the relationship between representation bias and intersectionality in real data. 


It remains to calibrate our damage metric (Definition \ref{def:kld_damage}) on the unrepaired data as in our repair metric $\hat{E}$ (\ref{eq:E-hat}), perhaps via the cross-entropy of the repaired data relative to unrepaired data.

One interesting avenue for future work is regarding exploiting the early stopping on some $u,s$ subgroups to begin conducting repairs (\ref{eq:repair}) sooner for samples $x_{u,s}$ from these quenched processes. Especially under data-streaming paradigms with large volumes of data, this approach has the potential to smooth the computational overhead of the mode-change from learning to repair. Additionally, it remains to adapt this method to non-stationary distributions, i.e. those with some drift in the underlying data-generating process, since the stationarity assumption in this work is particularly restrictive. Existing notions of forgetting \cite{kulhavy1993general}, specifically the work by \cite{quinn2007learning} on Bayesian non-parametrics for non-stationary processes, could be adapted to match our data repair paradigm and generalise our approach to a much broader class of $\O{F}_{u,s}$.

\section{Conclusion}
In this work, we have presented a novel method for fairness correction which exploits a data-driven stopping rule to alleviate representation bias in data by ensuring all $u,s$ conditional distributions are learned completely through a non-parametric stopping rule.

We present a comprehensive study of the performance of our Bayesian approach to learning the underlying distribution of sequential data under different operating conditions and when applied to diverse data modalities to simulate real world fairness correction. Further, we demonstrate that our repair approach is able to repair data even with severe representation bias, where the minority class is observed fewer than one time in 20, yielding repair quality and damage which are invariant to this bias. We also compare our method to the SOTA geometric and distributional repair operations on simulated GMM data and the Adult Income dataset.

Our method consistently outperforms the SOTA where representation bias is present, demonstrating the versatility of this data-driven approach. With the recent deployment of the AI act, this is a key step towards the generalisability of fairness tools for machine learning applications.


\vskip 6mm
\noindent{\bf Acknowledgements}

\noindent
This work has received funding from the European Union’s Horizon Europe research and innovation programme under grant agreement No. 101070568. This work was also supported by Innovate UK under the Horizon Europe Guarantee; UKRI Reference Number: 10040569 (Human-Compatible Artificial Intelligence with Guarantees (AutoFair)).

\bibliographystyle{abbrv}
\bibliography{bibliography}

\end{document}